\documentclass[letterpaper, 10 pt, conference]{ieeeconf}  % Comment this line out if you need a4paper

\IEEEoverridecommandlockouts                              % This command is only needed if 
\usepackage{graphicx}
\usepackage[flushleft]{threeparttable}
\usepackage{multirow}
\usepackage{xcolor}
\usepackage{amsmath}
\usepackage{amssymb}
\usepackage{cite}
\usepackage{times}
\usepackage{tabularx}
\usepackage{booktabs}
\usepackage{bm}
\usepackage{array}
\usepackage{soul}
\usepackage{algorithm}
\usepackage{algpseudocode}
\usepackage{colortbl}

\usepackage{cuted}
\usepackage[font=small,labelfont=bf]{caption}
\usepackage[colorlinks,linkcolor=red,anchorcolor=blue,citecolor=green]{hyperref}

\definecolor{mygreen}{HTML}{A9D18E}
\definecolor{myblue}{HTML}{9DC3E6}
\definecolor{myorange}{HTML}{F4B183}
\definecolor{mypurple}{HTML}{CC99FF}

\renewcommand{\baselinestretch}{0.995}

\title{\LARGE \bf
Robot Learning from a Physical World Model
}

\author{
    Jiageng Mao\textsuperscript{$\blacktriangle,\bigstar$} \quad
    Sicheng He\textsuperscript{$\bigstar$} \quad
    Hao-Ning Wu\textsuperscript{$\blacktriangle$} \quad
    Yang You\textsuperscript{$\clubsuit$} \quad 
    Shuyang Sun\textsuperscript{$\blacktriangle$} \quad
    Zhicheng Wang\textsuperscript{$\blacktriangle$} \quad \\
    Yanan Bao\textsuperscript{$\blacktriangle$} \quad
    Huizhong Chen\textsuperscript{$\blacktriangle$} \quad 
    Leonidas Guibas\textsuperscript{$\blacktriangle,\clubsuit$} \quad
    Vitor Guizilini\textsuperscript{$\diamondsuit$} \quad
    Howard Zhou\textsuperscript{$\blacktriangle$} \quad
    Yue Wang\textsuperscript{$\bigstar$} \\[0.5em]
    \textsuperscript{$\blacktriangle$}Google DeepMind \quad
    \textsuperscript{$\bigstar$}USC \quad 
    \textsuperscript{$\clubsuit$}Stanford \quad
    \textsuperscript{$\diamondsuit$}Toyota Research Institute
}

\begin{document}

\maketitle
% ===== Teaser right under the title (full width, non-floating) =====
\begin{strip}
\centering
% Use \textwidth to span both columns; adjust vspace as you like.
\vspace{-4mm}
\includegraphics[width=\textwidth]{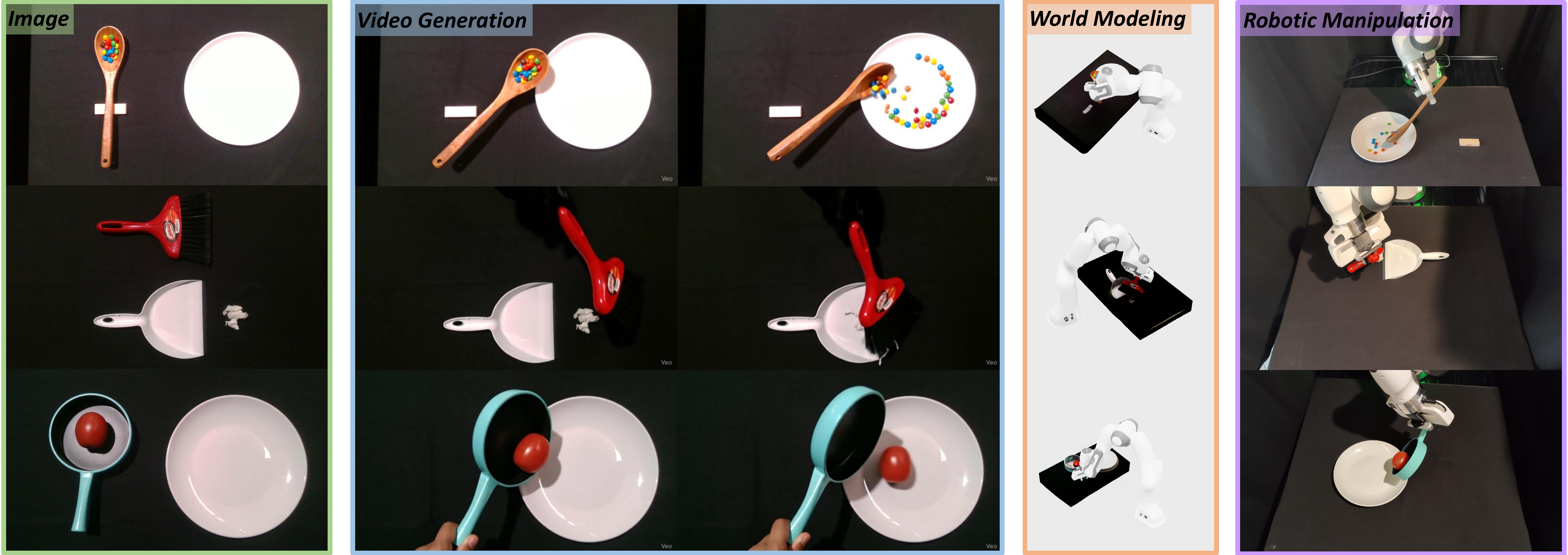}
\vspace{-3mm}
% \captionof{figure}{We present PhysWorld, a framework for robot learning from video generation. Given an image and a task prompt as inputs \textbf{\textcolor{mygreen}{(cololum \#1)}}, our method generates a task-conditioned video \textbf{\textcolor{myblue}{(cololum \#2)}} and models the underlying physical world for grounding generative video demonstrations into physical robot actions \textbf{\textcolor{myorange}{(cololum \#3)}}, enabling zero-shot robotic manipulation in the real world \textbf{\textcolor{mypurple}{(cololum \#4)}}.}
\captionof{figure}{\textbf{PhysWorld: a framework for robot learning from video generation.} Given an image and a task prompt as inputs \textbf{\textcolor{mygreen}{(column \#1)}}, our method generates a task-conditioned video \textbf{\textcolor{myblue}{(column \#2)}} and reconstructs the underlying physical world to ground generated visual demonstrations into physically feasible robot actions \textbf{\textcolor{myorange}{(column \#3)}}, enabling zero-shot robotic manipulation in the real world \textbf{\textcolor{mypurple}{(column \#4)}}.}
\vspace{2mm}
\end{strip}
% ================================================================

% \begin{figure*}[!t]
% \vspace{-3mm}
% \centering
% \includegraphics[width=0.99\linewidth]{figs/teaser.png}
% \vspace{-5mm}
% \caption{PhysWorld.}
% \label{fig/teaser}
% \vspace{-3mm}
% \end{figure*}

\thispagestyle{empty}
\pagestyle{empty}

%%%%%%%%%%%%%%%%%%%%%%%%%%%%%%%%%%%%%%%%%%%%%%%%%%%%%%%%%%%%%%%%%%%%%%%%%%%%%%%%
\begin{abstract}

We introduce PhysWorld, a framework that enables robot learning from video generation through physical world modeling. Recent video generation models can synthesize photorealistic visual demonstrations from language commands and images, offering a powerful yet underexplored source of training signals for robotics. However, directly retargeting pixel motions from generated videos to robots neglects physics, often resulting in inaccurate manipulations. PhysWorld addresses this limitation by coupling video generation with physical world reconstruction. Given a single image and a task command, our method generates task-conditioned videos and reconstructs the underlying physical world from the videos, and the generated video motions are grounded into physically accurate actions through object-centric residual reinforcement learning with the physical world model. This synergy transforms implicit visual guidance into physically executable robotic trajectories, eliminating the need for real robot data collection and enabling zero-shot generalizable robotic manipulation. Experiments on diverse real-world tasks demonstrate that PhysWorld substantially improves manipulation accuracy compared to previous approaches. Visit \href{https://pointscoder.github.io/PhysWorld_Web/}{the project webpage} for details. 
% \yue{should we put a website?}.\vitor{a website with videos would be great!}

\end{abstract}

\section{Introduction}
% \yue{Introduction reads great! It can be further improved by making it sound more profound. How does this method shape future robot learning? How does this method solve data problems for robotics? How does this method translate any AIGC to robotic ready data? I don't have a good solution here but just want to make sure the introduction sound more intriguing. I'll take another look on Monday}
Recent advances in generative models enable the synthesis of photorealistic videos directly from images and language instructions. Trained on large-scale Internet data, video generation models exhibit strong generalization across diverse scenarios. For robotics, such models offer a powerful source of visual guidance for manipulation. Given a robot’s observation and a task instruction, a video generator can produce a demonstration that depicts task completion. These generated videos inherently capture object dynamics and embodiment motions, which can be leveraged to learn generalizable robotic manipulation policies.

Despite the great promise of video generation, translating generated pixel motions into executable robotic actions remains highly challenging. Previous works~\cite{dreamitate,thisandthat,predictive_inverse_dynamics,dreamgen,unipi} learn inverse dynamics or policy models to align generated video frames with real robotic actions.
Yet, such methods generally rely on large-scale real-world demonstrations for alignment, while collecting them at scale is costly and labor-intensive.
% However, they require a sufficient amount of real-world robotic demonstrations for alignments, and collecting them at scale is time and labor intensive. 
Other methods~\cite{avdc,gen2act,RIGVid} propose to extract robotic actions by directly following visual cues, \textit{e.g.} flows, sparse tracks, or object poses, from generated videos. Nevertheless, directly retargeting video motions to robots neglects underlying physical constraints, often leading to inaccurate manipulations.

We argue that the key bottleneck of bridging generated videos and robotic actions lies in physical feasibility. Video generation, despite its generalization power, only provides visual plausibility rather than physical accuracy for robotic tasks, whereas robots operating in the real world require physically accurate actions to interact with objects correctly. We tackle this dilemma by introducing a proxy physical world model built from generated videos. This world model provides realistic physical feedback, enabling scalable robot learning to imitate generated video motions in a physically consistent manner.

To this end, we propose PhysWorld, a framework for physically grounded robot learning from video generation. The core of PhysWorld lies in the synergy between physical world reconstruction and video generation: video generation provides pixel-level visual guidance for task execution, while the physical world model offers realistic feedback for learning from the generated visual guidance. In particular, given a single RGB-D image and a task prompt, our method first generates a task-conditioned video depicting how the task is completed visually. Next, we propose a novel method for constructing a physically interactable scene from the video. Finally, we introduce an object-centric residual reinforcement learning approach that bridges video generation and physical world reconstruction, producing physically accurate robotic actions. Our framework requires only a single RGB-D image and a language command, yet outputs executable actions that follow the instruction to complete the task. By explicitly modeling physics, PhysWorld eliminates the need for real-world data collection and achieves zero-shot generalizable robotic manipulation, while significantly improving manipulation accuracy over previous methods.

We evaluate PhysWorld on a diverse set of real-world robotic manipulation tasks. Experimental results show that combining video generation with physical world modeling yields substantial improvements in accuracy across all tasks. PhysWorld enables physically grounded and generalizable robotic manipulation, consistently outperforming existing approaches by a large margin. We will release code and project resources to facilitate further research.
\section{Related Works}

\textbf{Video generation for robotics.} Video generation~\cite{cogvideox, cosmos, tesseract} holds great promise for robotics. It has been explored for goal generation~\cite{generativeimage}, planning~\cite{thisandthat,videolanguageplanning}, dynamics learning~\cite{unisim, uva}, and policy learning~\cite{avdc,dreamgen,dreamitate,RIGVid,predictive_inverse_dynamics,gen2act,unipi}. To extract robotic actions from generated videos, several works~\cite{dreamitate,thisandthat,predictive_inverse_dynamics,dreamgen,unipi,unisim,uva} train action models from generated video frames using large amounts of real robotic demonstrations, but collecting such data is costly. In contrast, PhysWorld removes the need for real-world data collection and enables zero-shot robotic manipulation. Other approaches~\cite{avdc,gen2act,RIGVid} directly extract actions by following visual cues from generated videos, such as optical flows~\cite{avdc}, sparse tracks~\cite{gen2act}, or object poses~\cite{RIGVid}. However, pixel-level imitation neglects physical plausibility and often results in inaccurate real-world manipulation. PhysWorld instead introduces a proxy physical world model, allowing agents to imitate generated video motions with physical feedback, thereby improving the accuracy and feasibility of real-world robotic manipulation.

\textbf{Robot learning from videos.} Videos contain rich motion and task information that can be leveraged for training robotic policies. Researchers tackle this problem by learning transferable representations~\cite{rep-bc,rep-humantorobot,rep-mime,rep-mimicplay,rep-mismatch,rep-structured,rep-suboptim,rep-third,rep-uniskill,rep-vid2robot,rep-xskill}, tracking embodiment-agnostic motion representations~\cite{track-atp,track-ditto,track-flow,track-handme,track-learnbywatch,track-motiontrack,track-phantom,track-r+x,track-robotseerobotdo,track-spot,track-track2act,track-vision,track-you,track-zeroshot}, real-to-sim~\cite{r2s-mimicgen,r2s-oneshot,r2s-video2policy,r2s-xsim}, or reinforcement learning~\cite{rl-avid,rl-human2sim2robot}. PhysWorld shares similar insights with other works on object pose tracking~\cite{track-spot, RIGVid}, real-to-sim reconstruction~\cite{r2s-video2policy,r2s-xsim,rl-human2sim2robot,videomimic}, and reinforcement learning~\cite{r2s-video2policy,rl-human2sim2robot}. However, these methods generally rely on \textit{ad-hoc} laboratory settings for real-to-sim reconstruction and human demonstration collection, which limits their generalization to in-the-wild generated videos that often contain motion blur or visual hallucinations. In contrast, PhysWorld only requires a single generated video for real-to-sim reconstruction and can effectively learn physically accurate robotic actions from generated videos.

\begin{figure*}[t]
    \centering
    \includegraphics[width=\textwidth]{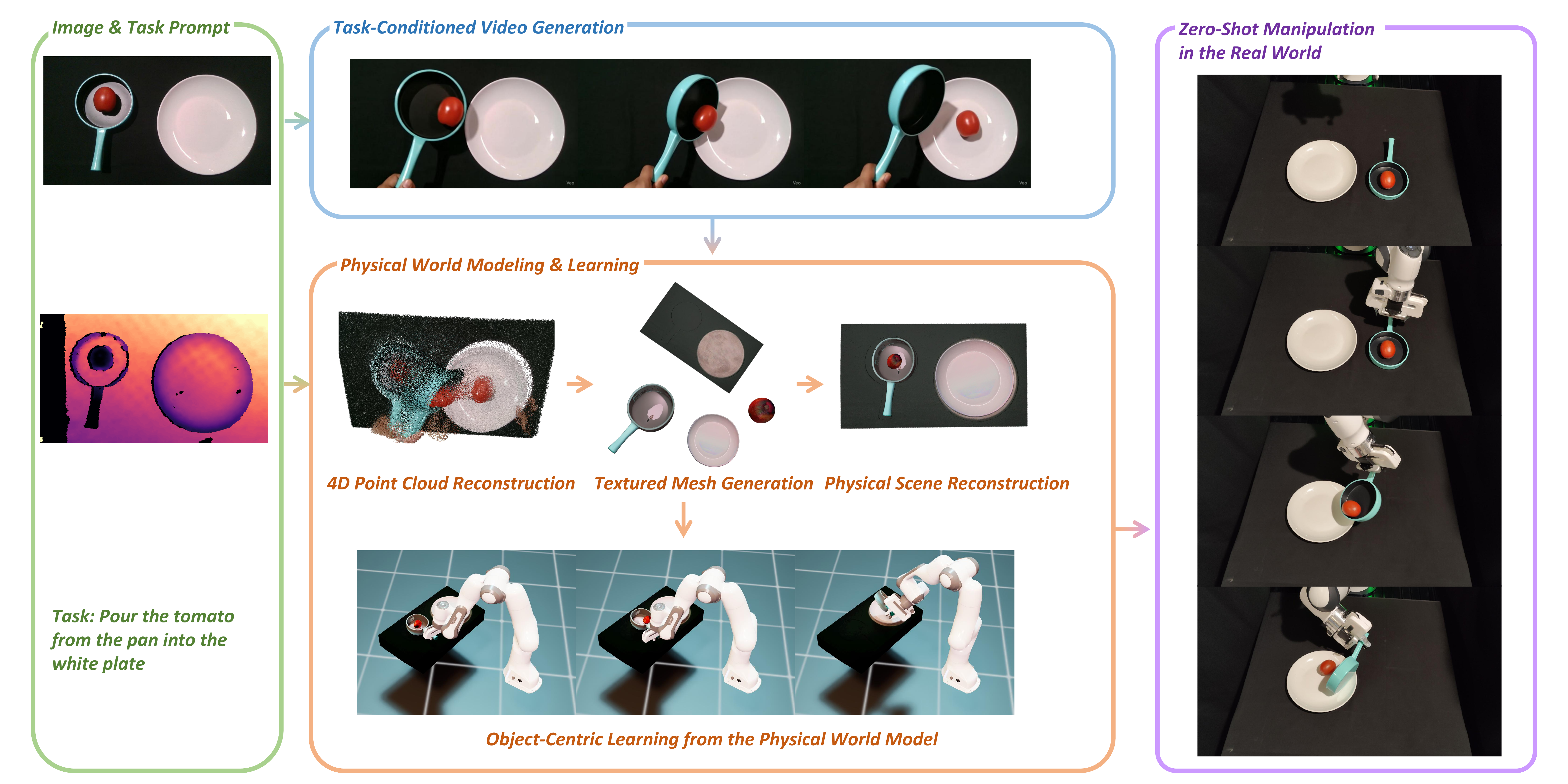}
    \vspace{-3mm}
    \caption{\textbf{PhysWorld pipeline.} 
    Given an RGB-D image and a task prompt, our framework (i) generates a task-conditioned video, 
    (ii) reconstructs a geometry-aligned 4D representation from the generated video, 
    (iii) generates textured object and background meshes, 
    (iv) assembles them into a physically interactable scene through property estimation, gravity alignment, and collision optimization, 
    (v) learns object-centric residual RL policies that transform visual demonstrations into feasible robotic actions,
    and (vi) deploys to the real world.}
    \label{fig:method}
    \vspace{-3mm}
\end{figure*}

\textbf{Real-to-sim-to-real.} Real-to-sim-to-real methods reconstruct a physical scene from observations and embed it in simulators for policy learning. To obtain complete object and scene textured meshes~\cite{rola} or Gaussian splats~\cite{3DGS}, prior works~\cite{real2sim-acdc,real2sim-grs,real2sim-physgs,real2sim-pulkit,real2sim-r3sim,real2sim-rebot,real2sim-robogs,real2sim-scalable,real2sim-simplerenv,real2sim-superlinear,real2sim-vlm,real2sim-vrrobo,realsim-robogsim} require dedicated multi-view captures for reconstruction, making them difficult to apply to monocular generated videos. In contrast, PhysWorld leverages generative priors to model physical scenes from a single-view video, enabling physical world modeling directly from generated videos without additional multi-view capture.

\section{Method}

We study the problem of open-world robotic manipulation. Our system takes as input an RGB-D image and a language-based task command, and outputs physically feasible robotic actions to complete the task. At its core, our approach unifies video generation and physical world modeling: video generation provides pixel-level visual guidance for task execution, while the physical world model offers realistic feedback for learning from the generated visual guidance. In Section~\ref{sec-3.1}, we describe how to model the physical world from generated videos, and in Section~\ref{sec-3.2}, we detail how to learn robotic actions from the physical world model.

\subsection{Physical World Modeling from Video Generation} \label{sec-3.1}

Video generation models trained on Internet data have demonstrated remarkable capability in generating visual demonstrations across diverse tasks and scenarios. However, these generated demonstrations only provide pixel-level guidance for task completion, while robots operate in the 3D space and are under physical constraints. To bridge this gap, we propose to first model the physical world from generated videos, transforming pixel-level guidance into physically grounded representations that can be executed by robots as accurate and feasible actions. Such a transformation is non-trivial, as generated videos provide only partial observations of the physical world and often contain visual artifacts. In this paper, we introduce a novel method that effectively tackles this problem with generative priors. Specifically, given a generated video, we first estimate a 4D spatio-temporal representation. We then generate textured meshes for objects and the background, endow them with physical properties, and align them with the 4D representation to construct the physical scene. Finally, we extract 4D motions from the video as targets for policy learning. The details of each step are presented in the following sections.

\textbf{Video generation.} Our method supports a variety of video generation models~\cite{tesseract,cogvideox,cosmos,veo3}, as long as they are image-to-video models with text control. Given an input image $I_{0}$ and a task command, a video generation model produces $T$ future frames $\{I_1, \dots, I_{T}\}$ demonstrating how the task will be completed. In this work, we primarily use Veo3~\cite{veo3} for video generation due to its high output quality, while additional models are evaluated in the ablation studies.

\textbf{Geometry-aligned 4D reconstruction.} Generated videos provide pixel-level demonstrations, and converting them into 4D spatio-temporal representations is necessary for robots that operate in the physical world. To obtain an accurate structure and motion estimate from videos, we initialize the dynamic scene reconstruction with MegaSaM~\cite{megasam}, which produces a temporally consistent depth estimate $\{D^{\prime}_0, \dots, D^{\prime}_{T}\}$ for each frame. However, MegaSaM's estimates are not well aligned with real-world metric scales. To address this, we leverage the real-world captured depth image $D_{0}$ to calibrate the outputs. Specifically, we solve for a global scale and shift $(\alpha, \beta)$ such that $\alpha D^{\prime}_0 + \beta \approx D_0$ over all valid pixels, by minimizing a robust regression objective:
\begin{equation}
\min_{\alpha, \beta} \; \sum_{p \in \Omega} w_p \, \big(\alpha \, D^{\prime}_0(p) + \beta - D_0(p)\big)^2 ,
\label{eq:1}
\end{equation}
where $\Omega$ denotes the set of valid pixels and $w_p$ are Huber weights that downweight outliers. 
The calibrated parameters $(\alpha, \beta)$ are then applied to all frames $\{D^{\prime}_t\}_{t=0}^T$, producing metric-aligned depth maps  $\{D_t\}_{t=0}^T$ that enable consistent 4D spatio-temporal reconstruction of the scene geometry. With known camera parameters, we can also obtain dynamic point clouds $\{P_t\}_{t=0}^T$ through un-projection.

\textbf{Textured mesh generation.} 4D reconstruction provides structures and motions from generated videos, but the depth or point cloud representation is not directly usable for physics simulation. Mesh is the standard geometry representation in simulators. Previous real-to-sim methods typically rely on pipelines such as Polycam or BundleSDF~\cite{bundlesdf} to reconstruct meshes from complete multi-view scans. However, these pipelines are unsuitable for generated monocular videos, where objects and scenes are only partially visible. To address this challenge, we propose a generative approach for recovering complete object and background meshes. 

Given the first image and its point cloud geometry ${I_0, P_0}$, we first separate objects from the background in $I_0$. The object pixels are removed, and the missing regions are filled using masked image inpainting~\cite{objectclear}, resulting in completed background imagery $I^{b}$ and individual object crops $I^{o}$. For each object, we apply an image-to-3D generator~\cite{trellis} to $I^{o}$, producing a canonical textured mesh $M^o$.

For background reconstruction, we require geometry $P^b$ corresponding to the completed background image $I^{b}$. This means inferring geometry in the regions originally occluded by objects. We address this with an object-on-ground assumption: objects are supported by the background, so their occluded regions are either planar supporting surfaces or extend to infinity (bounded by scene limits). Concretely, we cast camera rays through occluded pixels and compute their nearest intersections with either the supporting plane or scene boundaries, thereby filling in $P^b$ with consistent geometry. With ${I^b, P^b}$, we then reconstruct the background mesh $M^b$ via height-map triangulation and apply $I^b$ as the texture.

Finally, object and background meshes $\{M^o, M^b\}$ are assembled into a complete scene by aligning and resizing them to match the observed point cloud $P_0$ through registration.

% \begin{figure}[t]
%     \centering
%     \includegraphics[width=0.48\textwidth]{figs/viz_recon.png}
%     \vspace{-3mm}
%     \caption{\textbf{Qualitative evaluation of physical scene modeling.}}
%     \label{fig:recon}
%     \vspace{-3mm}
% \end{figure}

\begin{figure*}[t]
    \centering
    \includegraphics[width=0.95\textwidth]{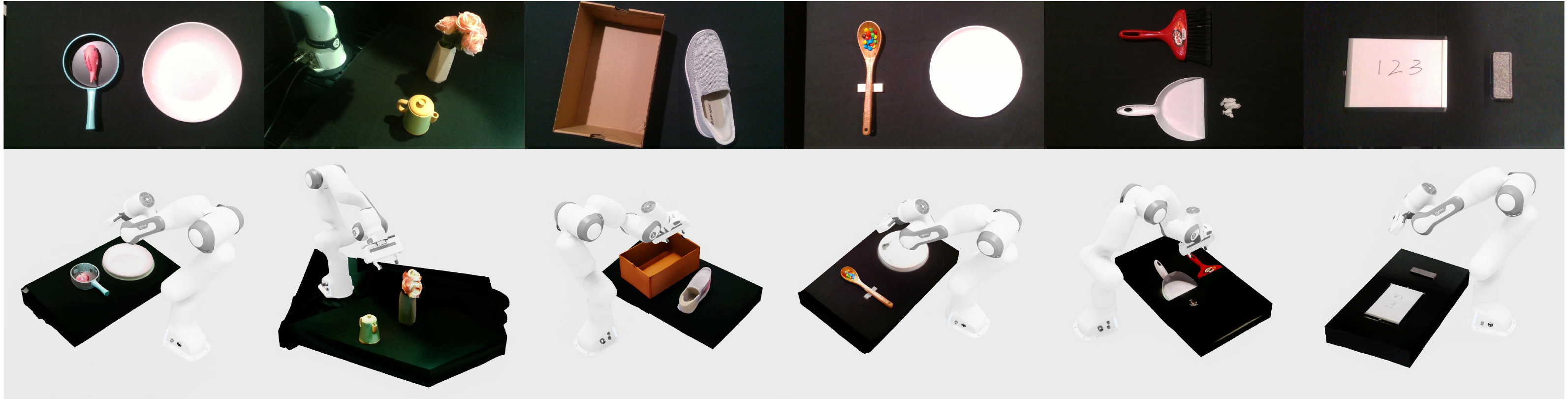}
    % \vspace{-3mm}
    \caption{\textbf{Qualitative evaluation of physical scene modeling from generated videos.}}
    \label{fig:recon}
    \vspace{-6mm}
\end{figure*}

\textbf{Physical scene reconstruction and alignment.} From the generated videos, we obtain decomposed scene meshes $\{M^o, M^b\}$. To make these meshes physically interactable, three additional steps are required: physical property estimation, gravity alignment, and collision optimization. 

Physical property estimation assigns appropriate physical parameters, such as mass and friction coefficients, to scene components. Inspired by~\cite{physproperty}, we leverage commonsense knowledge from vision-language models (VLMs) to estimate these properties. Specifically, we query a VLM with the object category to obtain typical physical parameters, and assign the predicted values to each object and the background for subsequent physical simulation.

Gravity alignment is to transform $\{M^o, M^b\}$ from camera to world frame so that the scene is consistent with the world gravity axis, which is crucial for physically plausible simulation. 
We estimate the ground plane normal $\mathbf{n}$ from segmented plane points using RANSAC, and compute the minimal rotation that aligns $\mathbf{n}$ with the world up axis $\mathbf{e}_z$: 
\begin{equation}
\mathbf{R}_{\text{grav}} = \exp\!\big([\mathbf{u}]_\times \,\theta\big), 
\theta = \arccos(\mathbf{n}^\top \mathbf{e}_z), 
\mathbf{u} = \frac{\mathbf{n} \times \mathbf{e}_z}{\|\mathbf{n} \times \mathbf{e}_z\|},
\end{equation}
where $[\mathbf{u}]_\times$ is the skew-symmetric matrix of $\mathbf{u}$. 
Applying $\mathbf{R}_{\text{grav}}$ to all meshes aligns the scene with gravity in the world frame for subsequent physical simulation.
%%%

Collision optimization is to optimize the placement of each object with respect to the background mesh so that all objects maintain a minimum clearance to avoid initial collisions. We voxelize the background mesh into a signed distance field (SDF) $\phi_{\text{bg}}$. 
For each object $M^o$, let $V_o=\{v_{o,1},\dots,v_{o,N_o}\}$ be its mesh vertices. 
We introduce a vertical translation $\tau_o$ along the gravity-opposing axis and solve
\begin{equation}
\min_{\{\tau_o\}} \; \sum_{o} \frac{1}{N_o}\sum_{i=1}^{N_o}
\Big[\max\!\big(0,\,-\phi_{\text{bg}}(v_{o,i}+\tau_o\,\mathbf{e}_z)\big)\Big]^2 ,
\label{eq:collision_opt}
\end{equation}
where $\mathbf{e}_z$ is the unit z-axis. 
This objective penalizes penetrations, i.e., negative SDF values, and is minimized by gradient descent using Adam with gradient clipping and early stopping. 
% After optimization, we enforce a clearance margin $\varepsilon$ by measuring
% \begin{equation}
% d_{\min}(o)=\min_{i}\,\phi_{\text{bg}}(v_{o,i}+\tau_o\mathbf{e}_z),
% \end{equation}
% and if $d_{\min}(o)<\varepsilon$, we apply an additional lift $\Delta z=\varepsilon-d_{\min}(o)$. 
This procedure ensures that all objects are adjusted relative to the background so that no initial collisions occur and a consistent clearance is preserved for simulation.

Finally, we obtain a physically interactable digital twin from the generated video. 
This physical model is essential for the subsequent learning process, as it provides the physically grounded feedback required to transform visual demonstrations into executable robotic actions.

\subsection{Object-Centric Learning from the Physical World Model} \label{sec-3.2}

With the physical world model established, the core step is to learn a robotic policy that can follow the generated video demonstrations. Video generation produces two types of motion: embodiment motion and object motion. Prior methods~\cite{rl-human2sim2robot} primarily retarget embodiment motions, but this often incurs high errors due to inaccurate motion transfer. The issue is further exacerbated for generated videos, which frequently contain hallucinated robots or human hands. In contrast, object motions are less prone to such artifacts and provide clearer visual guidance for task execution. Motivated by this, we focus on object-centric learning and introduce a residual reinforcement learning approach that tracks object motions under physical constraints.

\textbf{Learning targets.} Transforming generated visual demonstrations into 4D spatio-temporal learning objectives is necessary for training robotic policies. The commonly used learning objectives are optical flow~\cite{avdc}, object tracks~\cite{gen2act}, and object poses~\cite{RIGVid}. In this paper, we adopt object poses as tracking targets, since object pose estimation is generally more robust than other motion representations. Our framework also supports other forms of motion supervision, which we leave for future exploration. 
% Given the estimated 4D scene representations $\{D_t\}_{t=0}^T$ and $\{P_t\}_{t=0}^T$, together with object meshes $M^o$, 
% we use FoundationPose~\cite{foundationpose} to recover per-frame object poses $\{\mathbf{T}^o_t \in SE(3)\}_{t=0}^T$, which describe desired object positions in video demonstrations. 
% These trajectories $\{\mathbf{T}^o_t\}$ are incorporated as supervisions for policy learning, enabling the robot to track object motions in generated videos.
Given the estimated 4D scene representations $\{D_t\}_{t=0}^T$ and $\{P_t\}_{t=0}^T$, together with object meshes $M^o$, 
we use FoundationPose~\cite{foundationpose} to recover per-frame object poses
\begin{equation}
\{\mathbf{x}^o_t = [\mathbf{p}^o_t,\;\mathbf{q}^o_t]\}_{t=0}^T,
\end{equation}
where $\mathbf{p}^o_t \in \mathbb{R}^3$ is the object position and $\mathbf{q}^o_t \in \mathbb{R}^4$ is its orientation quaternion.  
These object pose trajectories $\{\mathbf{x}^o_t\}$ are incorporated as supervision for policy learning, enabling the robot to track object motions in generated videos.

\begin{figure*}[t]
    \centering
    \includegraphics[width=\textwidth]{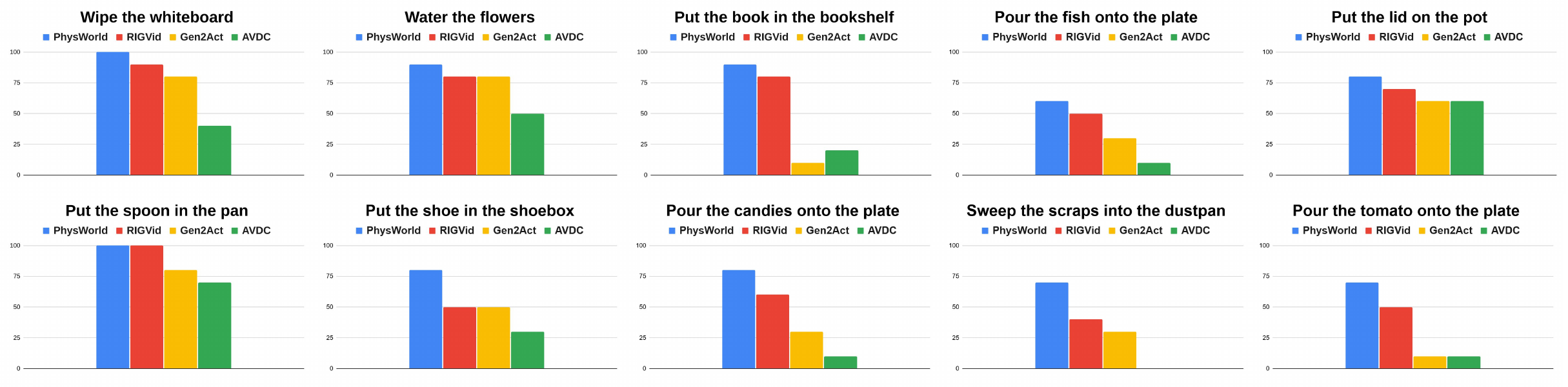}
    \vspace{-5mm}
    \caption{\textbf{Quantitative evaluation of PhysWorld on real-world manipulation tasks.}} 
    \label{fig:bench}
    \vspace{-4mm}
\end{figure*}

\textbf{Residual reinforcement learning.} 
A straightforward approach~\cite{RIGVid} to tracking object poses is to combine a grasping model~\cite{anygrasp} for object pickup with a motion planner~\cite{curobo} for subsequent placement. 
However, this strategy often struggles in complex manipulation tasks: grasping itself is prone to failure, and motion planning can also fail when initialized from improper poses. 
As a result, completing a task may require repeated grasping and planning attempts, leading to inefficiency and reduced reliability. 
Reinforcement learning is a promising alternative that can learn robust policies from physical feedback, but it often requires carefully designed rewards and long training time to converge. 
To address this, we propose a residual reinforcement learning method that combines the merits of both paradigms: grasping and motion planning provide baseline actions that narrow the search space, while an RL policy learns residual corrections on top of the baseline, enabling robust adaptation under feedback from the physical world model. 
Formally, given observation $\mathbf{o}_t$, the executed action is
\begin{equation}
\mathbf{a}_t = \mathbf{a}^{\text{base}}_t + \pi_\theta(\mathbf{o}_t),
\label{eq:residual_rl}
\end{equation}
where $\mathbf{a}^{\text{base}}_t$ is the baseline action from grasping and planning as in~\cite{RIGVid}, and $\pi_\theta(\mathbf{o}_t)$ is the residual policy that learns corrective adjustments. This residual formulation accelerates policy learning and improves robustness by leveraging feedback from the physical world model. Importantly, the success of the baseline itself is not required, since the learned residuals can rectify imperfect baseline actions to achieve task success.

\textbf{Observation and action space.} 
We adopt a state-based policy for efficient learning.
At each step $t$, the policy $\pi_\theta(\mathbf{o}_t)$ observes
\begin{equation}
\mathbf{o}_t = 
\big[\mathbf{x}^{\text{ee}}_t,\;
\mathbf{x}^{\text{obj}}_t,\;
\tau_t,\;
\mathbf{x}^o_t,\;
\mathbf{x}^{\text{grasp}},\;
d_{\text{pre}},\;
\mathbf{x}^{\text{base}}_t
\big],
\end{equation}
where $\mathbf{x}^{\text{ee}}_t$ and $\mathbf{x}^{\text{obj}}_t$ are the current end-effector and object poses, $\tau_t \in [0,1]$ is the normalized time index, $\mathbf{x}^o_t$ is the target object pose from the generated video. $\{\mathbf{x}^{\text{grasp}}, d_{\text{pre}}, \mathbf{x}^{\text{base}}_t\}$ are baseline actions from grasping and planning: $\mathbf{x}^{\text{grasp}}$ is a grasp proposal, $d_{\text{pre}}$ is a pre-grasp offset, and  
 $\mathbf{x}^{\text{base}}_t$ is the planned end-effector pose at time $t$, 
 
The policy outputs a residual action
$[\Delta \mathbf{p}_t,\;\boldsymbol{\omega}_t]$
consisting of a translation $\Delta\mathbf{p}_t \in \mathbb{R}^3$ and rotation $\boldsymbol{\omega}_t \in \mathbb{R}^3$. 
The executed command refines the baseline pose $\mathbf{x}^{\text{base}}_t$:
\begin{equation}
\mathbf{p}^{\text{cmd}}_t = \mathbf{p}^{\text{base}}_t + \Delta\mathbf{p}_t, 
\quad
\mathbf{q}^{\text{cmd}}_t = \exp([\boldsymbol{\omega}_t]_\times)\,\mathbf{q}^{\text{base}}_t,
\end{equation}
where $\mathbf{x}^{\text{base}}_t=[\mathbf{p}^{\text{base}}_t, \mathbf{q}^{\text{base}}_t]$, and $\mathbf{x}^{\text{cmd}}_t=[\mathbf{p}^{\text{cmd}}_t, \mathbf{q}^{\text{cmd}}_t]$ is the output end-effector pose command for robotic control. 

\textbf{Rewards.} We aim for simple rewards that can generalize to diverse tasks.
Specifically, an object tracking reward $r^{\text{trk}}_t$ encourages the robot to align the object with its target pose from the video:
\begin{equation}
r^{\text{trk}}_t = w_{\text{pos}}e^{-k_{\text{pos}}\|\mathbf{p}^{\text{obj}}_t-\mathbf{p}^o_t\|_2}
+ w_{\text{ori}}e^{-k_{\text{ori}}\|\mathbf{q}^{\text{obj}}_t-\mathbf{q}^o_t\|_2} .
\end{equation}
A grasp reward $r^{\text{grasp}}_t$ ensures a stable grasp and movement by penalizing excessive distance between the end-effector and the object when grasping and holding the object:
\begin{equation}
r^{\text{grasp}}_t = -w_{\text{grasp}}\,\mathbf{1}\!\left[\|\mathbf{p}^{\text{ee}}_t-\mathbf{p}^{\text{obj}}_t\|_2>\tau\right],
\end{equation}
where $\mathbf{p}^{\text{ee}}_t$ is the end-effector position,  
$\tau$ is a distance threshold, and $\mathbf{1}[\cdot]$ is the indicator function.

A planning reward $r^{\text{plan}}_t$ discourages infeasible actions by assigning a negative reward when inverse kinematics or motion planning fail.
We train the policy $\pi_\theta(\mathbf{o}_t)$ within the physical world model using the above reward terms, and adopt PPO~\cite{ppo} as the learning algorithm. Leveraging baseline actions significantly accelerates convergence, as the policy only needs to learn residual corrections. 

\section{Experiments}
% \yue{make some connections to intro and the key motivation}
% This section investigates three core questions:

% \textbf{Q1:} Does video generation enable more generalizable robotic manipulation? (Section \ref{Sec3-1})

% \textbf{Q2:} Does physical world modeling improve robustness in manipulation tasks? (Section \ref{Sec3-2}) 

% \textbf{Q3:} Does object-centric residual RL enhance policy effectiveness compared to other methods? (Section \ref{Sec3-3})

\begin{figure*}[t]
    \centering
    \includegraphics[width=\textwidth]{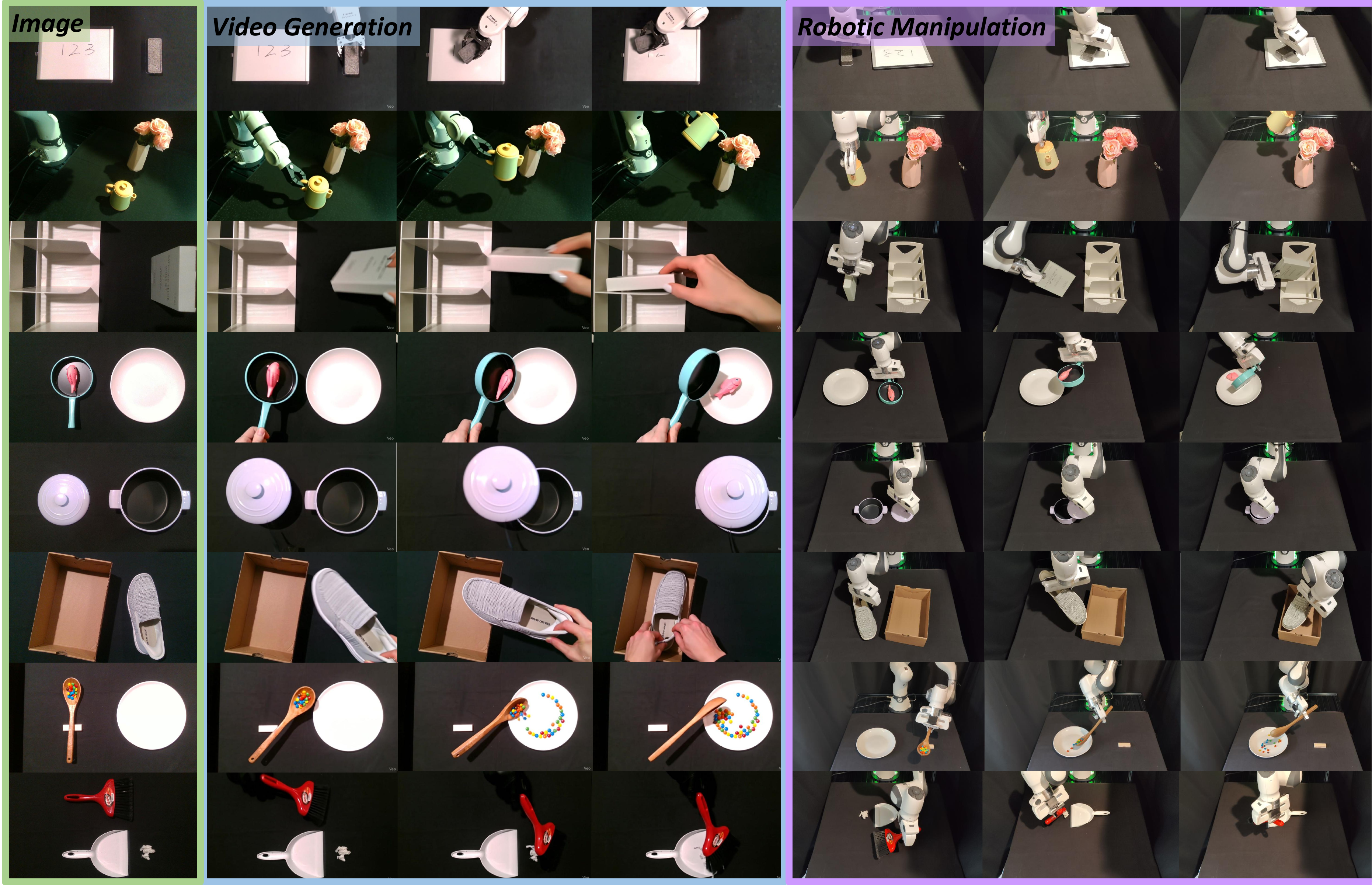}
    \vspace{-5mm}
    \caption{\textbf{Qualitative evaluation of PhysWorld on real-world manipulation tasks.}
    % \yue{shall we align the viewpoints of robot execution to those generated videos?}
    } 
    \label{fig:qual}
    \vspace{-3mm}
\end{figure*}

Our experiments aim to evaluate the efficacy of PhysWorld as a framework for unifying video generation and physical world modeling, quantify its ability to generalize without task-specific robot demonstrations, and analyze key design choices and limitations. To this end, we organize our study to answer the following empirical questions, in order:

\textbf{(Q1) Video Generation:} Does video generation enable more generalizable robotic manipulation? (Sec. \ref{Sec3-1})

\textbf{(Q2) World Modeling:} Does  physical world modeling improve robustness in manipulation tasks? (Sec. \ref{Sec3-2})

\textbf{(Q3) Learning:} Does object-centric residual RL enhance policy effectiveness compared to other methods? (Sec. \ref{Sec3-3})

\subsection{Video Generation Enables Generalizable Manipulation} \label{Sec3-1}

\textbf{Qualitative evaluation.} To answer whether video generation enables more generalizable robotic manipulation, we evaluate PhysWorld on a diverse set of real-world manipulation tasks, including: \emph{1. Wipe the whiteboard; 2. Water the flowers; 3. Put the book in the bookshelf; 4. Pour the fish from the pan onto the plate; 5. Put the lid on the pot; 6. Put the spoon in the pan; 7. Put the shoe in the shoebox; 8. Pour the candies from the spoon onto the plate; 9. Sweep the paper scraps into the dustpan; 10. Pour the tomato from the pan onto the plate}. A qualitative evaluation of PhysWorld on real-world manipulation tasks is shown in Figure~\ref{fig:qual}. The generated, task-conditioned videos provide rich, task-level visual guidance across diverse scenes, which our physical world model then grounds into executable actions, requiring \emph{no} additional robot data and enabling zero-shot robotic manipulation in the real world.

\textbf{Video generation quality.} To analyze how video generation quality impacts downstream manipulation, we compare 4 image-to-video models: Veo3~\cite{veo3}, Tesseract~\cite{tesseract}, CogVideoX1.5-5B~\cite{cogvideox}, and Cosmos-2B~\cite{cosmos}, on the same set of tasks. For each combination of model and task, we generate 10 videos and compute the fraction that are usable, \textit{i.e.}, those from which object poses can be recovered robustly. Table~\ref{fig:videomodels} reports the usable-video ratio across tasks. Veo3 achieves the highest overall ratio, and robotic data fine-tuning (\textit{e.g.}, Tesseract) tends to outperform generic generators. These results indicate that higher-quality, task-consistent video generation is necessary for reliable manipulation.

% \begin{figure}[t]
%     \centering
%     \includegraphics[width=0.48\textwidth]{figs/videogen.pdf}
%     \vspace{-3mm}
%     \caption{\textbf{Generation quality of different video generation models.} We measure the fraction of usable videos among all generated videos.
%     % \yue{what's the metric here?}\vitor{I'm curious as well, but this figure may go if we are running out of space}
%     }
%     \label{fig:videomodels}
%     \vspace{-3mm}
% \end{figure}

\begin{table}[t!]
  \renewcommand{\arraystretch}{1.1}
  \renewcommand\tabcolsep{30.0pt} 
  \footnotesize
  \caption{Generation quality of different video generation models. We measure the ratio of usable videos among all generated videos.}
    \vspace{-0.3cm}
    \begin{center}
    \begin{tabular}{l|c}
        \toprule[.03cm]
        Models & Usable ratio (\%) \\
        \midrule
        Veo3~\cite{veo3} & 70\%  \\
        Tesseract~\cite{tesseract} & 36\% \\
        CogVideoX1.5-5B~\cite{cogvideox} & 4\% \\
        Cosmos-2B~\cite{cosmos} & 2\% \\
        \toprule[.03cm]
    \end{tabular}
    \end{center} 
    \label{fig:videomodels}
    \vspace{-0.7cm}
\end{table}

\subsection{World Modeling Improves Manipulation Robustness} \label{Sec3-2}

\textbf{Physical scene reconstruction quality.} Figure~\ref{fig:recon} shows the reconstructed models from generated videos. Our method integrates geometry-aligned 4D reconstruction with generative priors to recover the underlying physical scenes from monocular inputs. The resulting scenes are geometry-consistent and physically interactable, providing reliable physical feedback for robot learning.

\textbf{Effectiveness of world modeling.} To evaluate the effectiveness of introducing physical world models, we evaluate 10 real-world manipulation tasks, each with 10 rollouts, and report the success rate of each task. We compare our method against 3 zero-shot methods without physical world modeling: (i) RIGVid~\cite{RIGVid}: it directly tracks object poses from generated videos and leverages off-the-shelf grasping models and motion planning for robotic control; (ii) Gen2Act~\cite{gen2act}: we adopt a modified version in~\cite{RIGVid}, which extracts sparse point tracks as tracking objectives; (iii) AVDC~\cite{avdc}: it leverages depth and optical flow estimation to represent object and embodiment motions. Figure~\ref{fig:bench} summarizes the quantitative comparison: PhysWorld attains the highest average success rate (82\%), significantly outperforming the second-best method~\cite{RIGVid} (67\%) by a large margin. This indicates that learning from a physical world model provides corrective feedback that reduces compounding errors from grasping and planning, especially in phases like picking, insertion, and pouring. Moreover, leveraging object poses as tracking targets significantly outperforms those using point tracks~\cite{gen2act} and optical flows~\cite{avdc}. This implies that object pose estimation provides more robust object motion signals from generated videos than point tracks and optical flow, which often suffer from drifting under occlusion and blurred motion.

\vspace{-2mm}
\begin{figure}[th]
    \centering
    \includegraphics[width=0.48\textwidth]{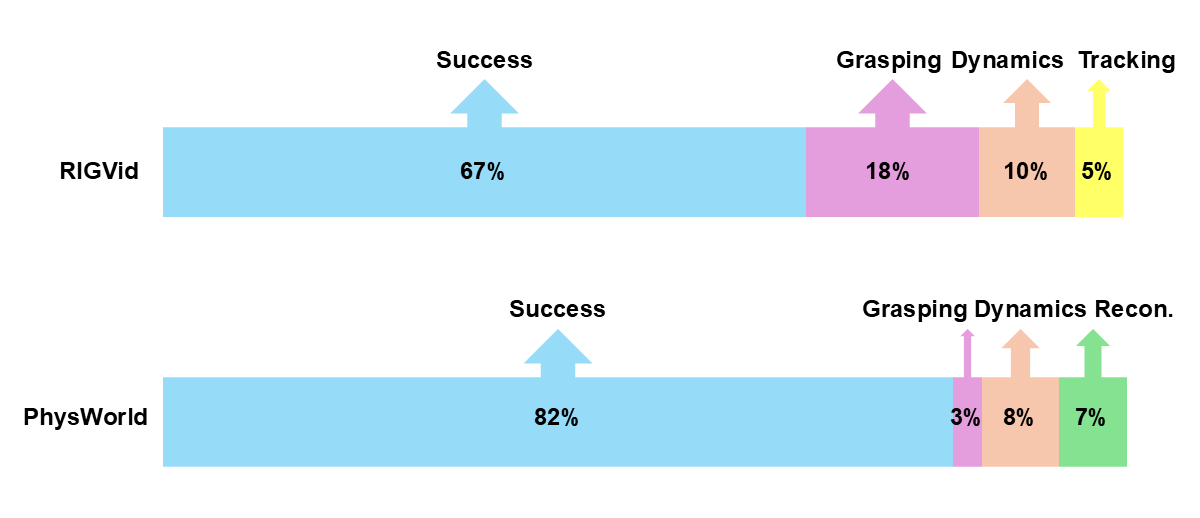}
    \vspace{-6mm}
    \caption{\textbf{Failure mode analysis.}}
    \label{fig:fail}
    \vspace{-3mm}
\end{figure}

\textbf{Failure mode analysis.} To further investigate where the performance gains come from, Figure~\ref{fig:fail} breaks failure cases into 4 categories: grasping, tracking, dynamics, and reconstruction. Comparing with~\cite{RIGVid}, introducing the physical world model substantially reduces grasping failures from 18\% to 3\% and eliminates tracking failures from 5\% to 0\%, which indicates the importance of physical feedback of world models. Our method introduces 7\% reconstruction errors. This is mainly because we reconstruct the physical scene from monocular, generated videos, and the completed geometry in occluded regions may be misaligned with real-world geometry. However, we argue that the problem can be mitigated by performing multiview reconstruction of the environment in advance.

\subsection{Object-Centric Learning Enhances Policy Effectiveness} \label{Sec3-3}

\textbf{Object-centric vs. embodiment-centric learning.} We compare two paradigms of learning from videos: (i) embodiment-centric learning, which reconstructs a human hand mesh and maps finger keypoints to the robot end-effector as robot movement trajectories, and (ii) object-centric learning, which trains policies to follow object motions. As shown in Table~\ref{tab:recon}, object-centric learning is remarkably stronger (\emph{Put the book in the bookshelf}: 90\% vs. 30\%; \emph{Put the shoe in the shoebox}: 80\% vs. 10\%). The main reason is that generated videos often hallucinate hands or exhibit inconsistent hand kinematics, whereas object motion is more stable and easier to estimate under occlusion. Object-centric learning therefore transfers more reliably to robots and aligns better with our physics-grounded training.

\begin{figure}[t]
    \centering
    \includegraphics[width=0.48\textwidth]{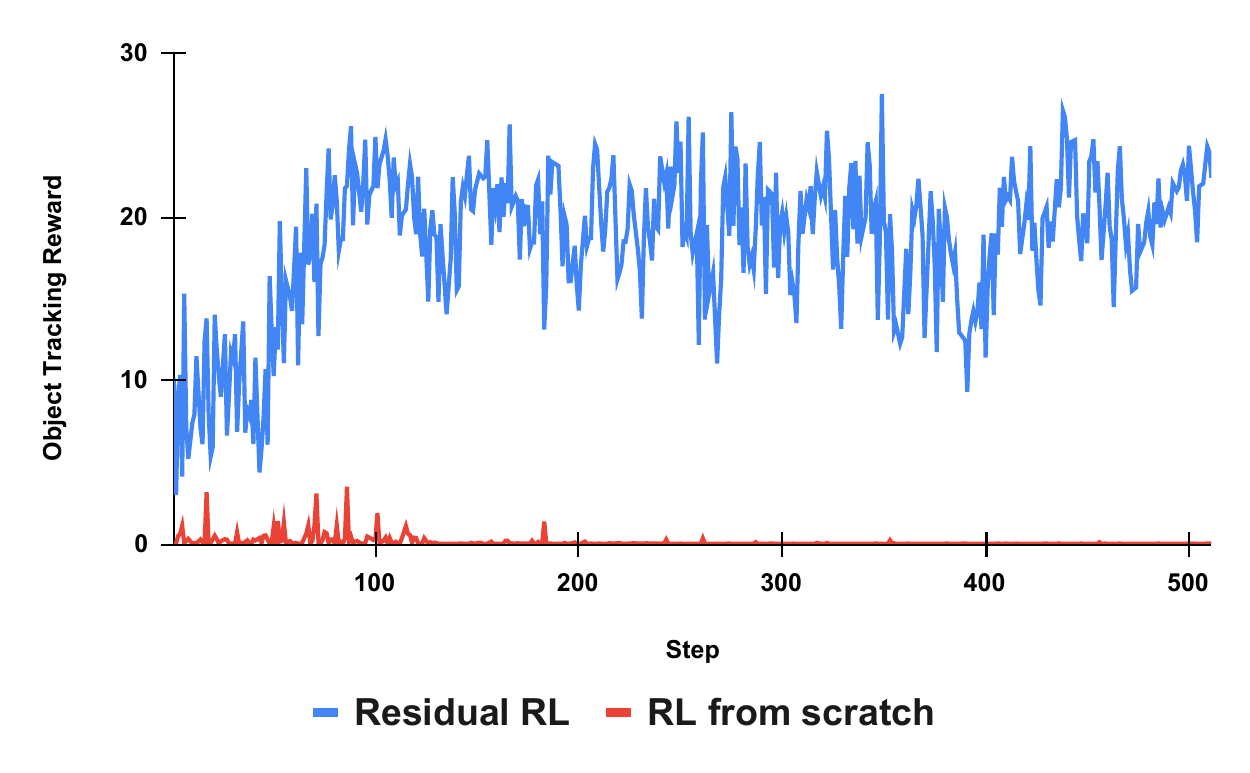}
    \vspace{-6mm}
    \caption{\textbf{Residual RL vs. RL from Scratch.}}
    \label{fig:rl}
    \vspace{-6mm}
\end{figure}

\textbf{Residual RL vs. RL from scratch.} We further compare residual RL with training a policy from scratch under the same physical world model and for the \emph{Pour the tomato from the pan onto the plate} task (see Figure~\ref{fig:rl}). Residual RL converges within a few hundred iterations and obtains higher object tracking rewards under the same budget. The baseline grasp-and-plan actions constrain exploration to a small, feasible neighborhood, while the world model provides corrective feedback that the residual policy uses to refine trajectories. In contrast, RL from scratch can also succeed~\cite{rep-humantorobot} but requires longer training time and more careful reward designs. Hence, residual RL with physical world models enables faster learning and improves the robustness of manipulation.
%These results support our design: pairing a physics-grounded simulator with residual corrections yields faster learning and more robust manipulation.

\begin{table}[th]
  \renewcommand{\arraystretch}{1.1}
  \renewcommand\tabcolsep{6.0pt} 
  \footnotesize
  \caption{Object-centric vs. embodiment-centric learning}
    \vspace{-0.3cm}
    \begin{center}
    \begin{tabular}{l|cc}
        \toprule[.03cm]
        Task & Embodiment-centric &  Object-centric \\
        \midrule
        Put the book in the bookshelf & 30\% & 90\% \\
        Put the shoe in the shoebox & 10\% & 80\% \\
        \toprule[.03cm]
    \end{tabular}
    \end{center} 
    \label{tab:recon}
    \vspace{-0.7cm}
\end{table}

\section{Conclusion}

We introduced PhysWorld, a framework that bridges video generation and robot learning through physical world modeling. By reconstructing a physically interactable scene from generated videos and learning object-centric residual RL policies, PhysWorld transforms generated visual demonstrations into physically feasible robotic actions, enabling zero-shot generalizable manipulation in the real world. Future work includes synthesizing physically accurate videos with this framework for training robotic video generation models.

\textbf{Limitations.} Physical world modeling is bounded by the fidelity of physical simulators and may introduce additional sim-to-real gaps. However, from the evidence in Figure~\ref{fig:bench}, we still believe in the necessity of introducing a world model to provide reliable physical feedback for more robust learning.

% \clearpage

\bibliographystyle{IEEEtran}
\bibliography{myref}

\end{document}